# Integration of Swin UNETR and statistical shape modeling for a semi-automated segmentation of the knee and biomechanical modeling of articular cartilage

Reza Kakavand[1], Mehrdad Palizi[2], Peyman Tahghighi[1], Reza Ahmadi[1], Neha Gianchandani[1], Samer Adeeb[2], Roberto Souza[3,4], W. Brent Edwards[1], Amin Komeili[1*]

[1]Department of Biomedical Engineering, Schulich School of Engineering, University of Calgary

[2]Civil and Environmental Engineering Department, Faculty of Engineering, University of Alberta

[3]Department of Electrical and Software Engineering, Schulich School of Engineering, University of Calgary

[4]Hotchkiss Brain Institute, Cumming School of Medicine, University of Calgary

**Corresponding author: Amin Komeili**

\* <amin.komeili@ucalgary.ca>

**Address: ICT243, 2500 University Drive NW, Calgary, AB, T2N 1N4**

# Abstract

Simulation studies like finite element (FE) modeling provide insight into knee joint mechanics without patient experimentation. Generic FE models represent biomechanical behavior of the tissue by overlooking variations in geometry, loading, and material properties of a population. On the other hand, subject-specific models include these specifics, resulting in enhanced predictive precision. However, creating such models is laborious and time-intensive. The present study aimed to enhance subject-specific knee joint FE modeling by incorporating a semi-automated segmentation algorithm. This segmentation was a 3D Swin UNETR for an initial segmentation of the femur and tibia, followed by a statistical shape model (SSM) adjustment to improve surface roughness and continuity. Five hundred and seven magnetic resonance images (MRIs) from the Osteoarthritis Initiative (OAI) database were used to build and validate the segmentation model. A semi-automated FE model was developed using this semi-automated segmentation. On the other hand, a manual FE model was developed through manual segmentation (i.e., the gold standard approach). Both FE models were subjected to gait loading. The predicted mechanical response of manual and semi-automated FE models were compared. In the result, our semi-automated segmentation achieved Dice similarity coefficient (DSC) over 98% for both femur and tibia. The mechanical results (max principal stress, max principal strain, fluid pressure, fibril strain, and contact area) showed no significant differences between the manual and semi-automated FE models, indicating the effectiveness of the proposed semi-automated segmentation in creating accurate knee joint FE models. ( https://data.mendeley.com/datasets/k5hdc9cz7w/1 ).

# 1 Introduction

Osteoarthritis (OA) is a degenerative joint condition with multifactorial etiology, including obesity, age, injury, and heredity [1]–[5]. OA affects 33.6% of people over 65 in the U.S. and 13.9% of those aged 25 and older, costing $22.6 billion annually solely in joint replacements [1]. The underlying mechanobiological mechanism(s) leading to the initiation and progression of OA are poorly understood.

Simulation studies, such as FE modeling, provide insight into the stresses, strains, and contact mechanics within the knee joint under physiologically relevant loading conditions [6]–[10]. Generic FE models typically predict the biomechanical behavior of the knee joint based on representative or aggregate data [11]. By using data from a cohort of subjects, researchers are able to create a simplified and standardized representation of the biological system or mechanical structure under investigation, which serves as a fundamental starting point for research studies. Alternatively, subject-specific FE models include personalized information that result in more accurate predictions [11]–[14]. The utility of subject-specific models is limited by their labor-intensive nature, time requirements, and lack of reproducibility [15], [16]. For the construction of subject-specific FE models, segmentation of computed tomography (CT) and magnetic resonance images (MRIs) has become the gold standard. However, manual segmentation is a labor-intensive task that may take several working days per individual [17], [18].

Convolutional neural network (CNN) and statistical shape models (SSM) have demonstrated the ability to accelerate segmentation operations of medical images. SSM involves a principal component analysis (PCA), performed on a training set of extracted subject geometries to determine its modes of spatial variation [19]–[22]. Different CNNs and SSMs have been successful at a variety of segmentation tasks [23]. Ambellan et al. [24] used CNN for 2D and 3D segmentation of knee tissues and implemented SSM to control regions with abnormal shape. The Dice similarity coefficient (DSC) values for the femur and tibia were 98.6% and 98.5% using data from Osteoarthritis Initiative (OAI). Paproki et al. [25] and Tack et al. [26] facilitated segmentation of menisci in healthy and OA knees using active shape modeling and SSM.

Deep Convolutional Neural Networks (DNNs), particularly the U-Net model, have demonstrated exceptional performance in medical image segmentation across different modalities and organs [27], [28]. However, CNN-based approaches often struggle to capture long-range dependencies due to their reliance on localized receptive fields.

Recent machine learning methods have significantly improved the segmentation of organs from biomedical images. For instance, UNETR is a neural network architecture that combines the strengths of UNet and transformer models for accurate image segmentation tasks [29]. Additionally, Swin UNETR [30] was developed for segmenting brain tumors from MRIs and demonstrated superior accuracy and efficiency in a variety of benchmarks. Swin UNETR combines the encoder from Swin transformers, a modified version of Vision Transformer (ViT), with the decoder inspired by 3D U-Net [31], [32]. Swin transformers, specialized for the visual domain, overcame the quadratic model complexity drawback of ViT by employing a shifted windowing scheme. The hierarchical structure of Swin transformers allows for modeling and combining image features at multiple scales, similar to CNNs. Furthermore, the linear computational complexity of Swin transformers enhances its efficiency for dense prediction tasks using high-resolution images [33]. Nevertheless, the outcome of the automatic segmentation methods requires manual correction to improve surface smoothness, fill holes, and correct abnormal morphologies. Therefore, 3D FE model preparation from biomedical images still requires significant human intervention and supervision. With the recent advancements in the field of medical image segmentation, the previous algorithms for knee joint cartilage segmentation could be revisited to enhance their accuracy and reproducibility for computational modeling. A gap exists between the existing advanced automated segmentatiom models and their implementation in biomechanical modeling. In addition, there seems to be a lack of publically available automated segmentation models suitable for subject-specific biomechanical modeling of knee joint.

This study aimed to develop an advanced semi-automated segmentation method for creating knee joint FE models. Therefore, the objectives of this study were: 1) to train a 3D Swin UNETR transformer and SSM for the semi-automated segmentation of distal femur and proximal tibia (the output of Swin UNETR which is an image needs to be manually converted to a 3D printing or computer-aided design format such as stl before using SSM) which is suitable for biomechanical modeling, and 2) to apply a gait loading condition to the segmentation for the assessment of the biomechanical modeling outcomes such as max principal stress, max principal strain, fluid pressure, fibril strain, and contact area . We have made our semi-automated models publicly accessible to support and facilitate biomechanical modeling and medical image segmentation efforts ( https://data.mendeley.com/datasets/k5hdc9cz7w/1 ).

## 2  Method

In this study, two FE models (manual FE model and semi-automated FE model) were generated. The geometry of the bones was the only difference between these two FE models. The geometry of the bones in the manual FE model was created using the manual segmentation of femur and tibia, whereas in the semi-automated FE model it was created using the semi-automated segmentation. The geometries for cartilages and menisci were manually segmented and added to both FE models. For the semi-automated segmentation, a 3D Swin UNETR transformer was used for the initial automatic segmentation of femur and tibia. The femur and tibia outcome from Swin UNETR was further adjusted with SSM to improve their surface quality in terms of surface roughness and hole filling. Bone surfaces (from both the manual and semi-automated FE models) were meshed using quadrilateral elements. The quadrilateral elements in the calcified region were extruded to the articular surface of the cartilage using hexahedral meshes. These hexahedral meshes represented cartilage in our FE models. Ligaments were modelled as bi-linear springs that could withstand tension but not compression (Figure 1). The predicted mechanical response of the manual and semi-automated FE models, including the cartilage contact mechanics and pore pressure, were compared. Specific details of these procedures are outlined below.

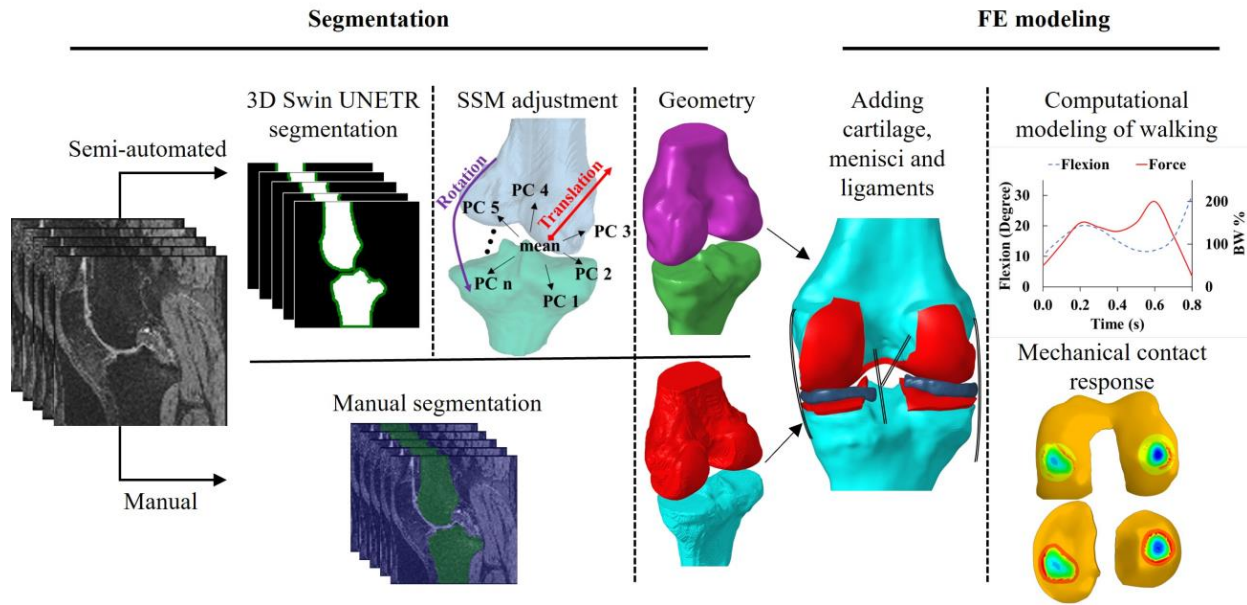

Figure 1. Workflow used in this study. Semi-automated and manual segmentations were developed from MRIs of knee joints from OAI database. The geometries from these segmentations were used to create semi-automated and manual FE models. Both FE models were subjected to gait loading and were compared for their mechanical response, including cartilage contact mechanics and pore pressure. PC is the principal components in SSM.

## 2.1 Data

A total of 507 MRIs (61.87±9.33 years; 29.27±4.52 BMI [kg/m$^2$]; 0.36×0.36×0.7 mm image resolution) were extracted from the OAI database. The masks for the femur, tibia and cartilages were performed by skilled users from the Zuse Institute Berlin [24]. The images include all OA grades, with a high tendency towards severe cases. To evaluate the performance of Swin UNETR and SSM models, we used 5 fold cross-validation (***Figure 1S in the supplementary material***). Each fold had 405 MRIs for training and 102 MRIs for testing. To evaluate the performance of FE models, we used 9 randomly selected samples from the test set since FE modeling of 102 samples is extremely time-consuming and currently infeasible (At least until the automation of the steps in FE modeling of knee joint such as meshing, material property assigning and loading is achieved).

## 2.2 Swin UNETR

The hierarchical structure of the Swin transformer allows for modeling and combining image features at multiple scales (like CNNs), and it maintains linear computational complexity in relation to image size [33]. The four output features extracted from the Swin transformer (indicated by red arrows) were fed into the 3D U-Net blocks to reconstruct an image with the same size as the input (Figure 2). The model yields a single output for each pixel. For 3D Swin UNETR, a patch size of 2, a window size of 7, and an initial feature size of 48 were used. The Swin transformer had four stages and utilized three 3D U-Net blocks for upsampling (Figure 2). The Swin UNETR was trained using the DSC and binary cross entropy focal loss.

Our proposed method was implemented in Python, and our deep learning models were implemented using the Pytroch library. All models were trained with a batch size of 8, using the Adam optimizer [34], [35] with an initial learning rate of 0.0001 and early stopping to avoid overfitting. The model was evaluated using 5-fold cross-validation (***Figure 1S in the supplementary material***). The original image size was 160,384,384. During training, each MRI was resized to $128 \times 128 \times 160$ pixels and cropped randomly to $96 \times 96 \times 96$ pixels. The resizing and cropping significanly improved the the generalization of the Swin UNETR model. We applied window center adjustment on MRIs as the preprocessing step to translate them to the range [0,1]. The data augmentation and Swin UNETR implementation were done using the Monai library (https://docs.monai.io/en/stable/). All the other variables were the default settings in the Monai. The Swin UNETR models and codes have been made publicly available, so researchers can use these models or customize the code for a different dataset to meet their needs ( https://data.mendeley.com/datasets/k5hdc9cz7w/1 ).

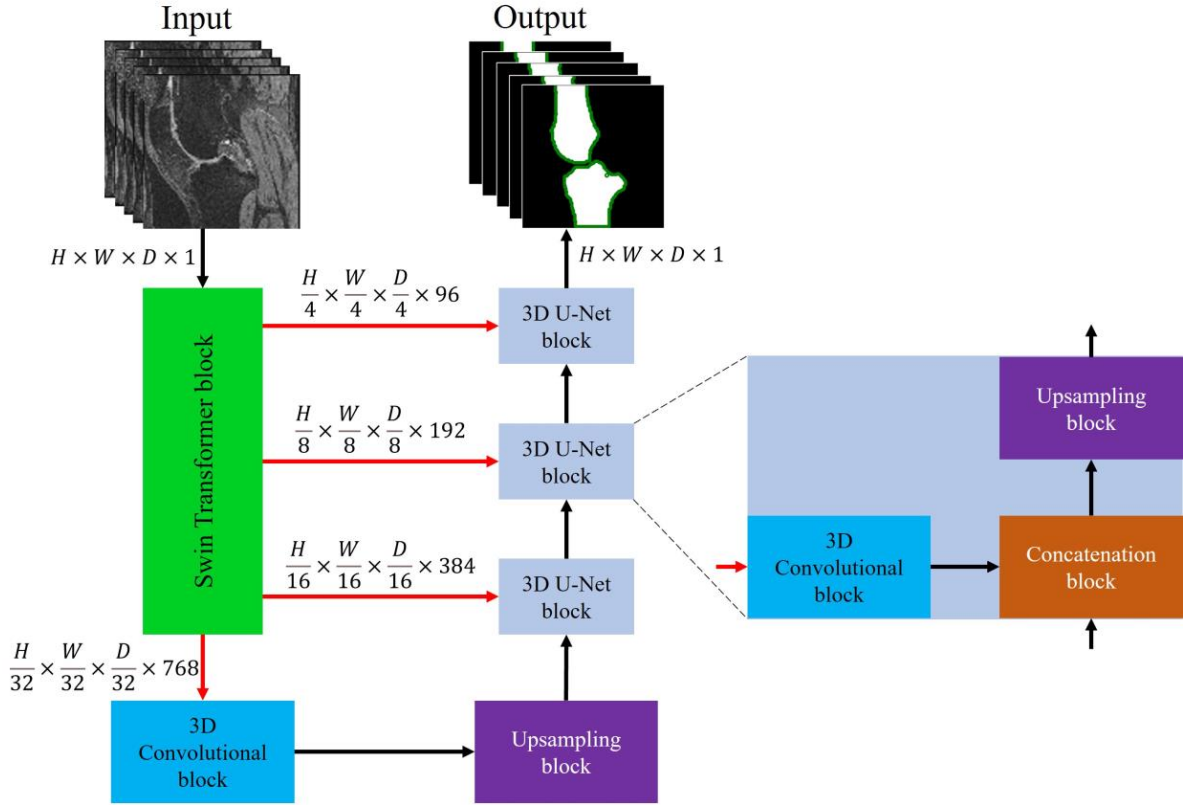

Figure 2. The structure of the employed Swin UNETR segmentation model. It was a combination of a Swin transformer as the encoder and 3D U-Net blocks as the decoder. Swin transformer extracted hierarchical representation from given MRIs and 3D U-Net utilized these representations to construct the segmentation mask. The model outputs a binary label for each pixel. Please note that for each femur and tibia, we developed a separate Swin UNETR segmentation model. Here, for the illustration, the outputs of femur and tibia were combined.

2.3  SSM

Using SSM, a shape may be defined as:

Shape = M + PC × b    (eq. 1)

where M is the mean of the points of the shape, PC ($PC_1$, $PC_2$, $PC_3$, …) is the principal components (the modes of variations of the points of the shape), and b is a vector of weights. To build an SSM for the femur and tibia, correspondence was first established between the samples, and then, Principal Component Analysis (PCA) was employed to model anatomical variation. The SSM was developed with a custom

Python script. Constructing correspondence between the samples included coarse and fine alignment steps using point-set representation. First, a manually segmented mesh belonging to one of the subjects was selected as the initial template. The template mesh for the femur and tibia was refined into meshes with optimized tessellation quality using the iso-parametrization method [36]. The mesh was then smoothed using the Taubin method [37]. A dense set of points with uniform distribution was sampled on each mesh (representing the femur or tibia for a subject or the template). To uniformly sample points on the meshes, the Poisson-disk point-set sampling method [38] was used to achieve well-distributed points, then the uniformization technique [39] was used to further homogenize the distance between the neighboring points.

For each subject, the combined point set for the femur and tibia was coarsely aligned to the template. The coarse alignment was performed using the centroid, the centroid size, and the principal axes of the combined point set [39]. Next, the template point set was matched on each sample. The matching process involved the rigid registration of the template to the sample, followed by a non-rigid registration. The coherent Point Drift (CPD) method [40] was used for the rigid and non-rigid registration tasks. After registering the template to all samples, the redundant rigid transformation within the samples was removed using the generalized Procrustes Analysis (GPA) [41]. The average shape for the femur and tibia was computed as the arithmetic mean of the point sets in correspondence (after applying GPA) and PCA was finally applied. To generate a mesh for each instance of the SSM, the deformation field between the template point set and the point set of the shape instance was decomposed into affine and non-rigid components using the Thin-Plate Spline (TPS) formulation [42], and the characterized transformation was applied to the vertices of the template (with high quality of tessellation). This process resulted in a representation of each sample (femur or tibia) with a deformed version of the template mesh with high-quality tessellation.

## 2.4 Cartilage Extrusion

The segmented femora and tibiae were meshed using quadrilateral elements. The femoral condyle and tibial plateau surfaces that share the calcified cartilage zone were mapped to the articular cartilage surface using 8 node solid elements, creating 5 layers of hexahedral elements (Figure 3). In this way, the variation of cartilage thickness over the joint was captured, and common nodes were defined at the interface of bone and cartilage. This aimed to accelerate computational time and improve the convergence rate.

Mesh sensitivity was performed with three different element sizes of 2, 1 and 0.5 mm (coarse, fine, and very fine). A 1% change in the contact area and average contact pressure between models were used as the convergence criterion to select the optimized mesh size. The difference in outputs between fine and very fine elements was less than 1%, indicating successful convergence at the fine resolutions, which was selected for further FE analyses.

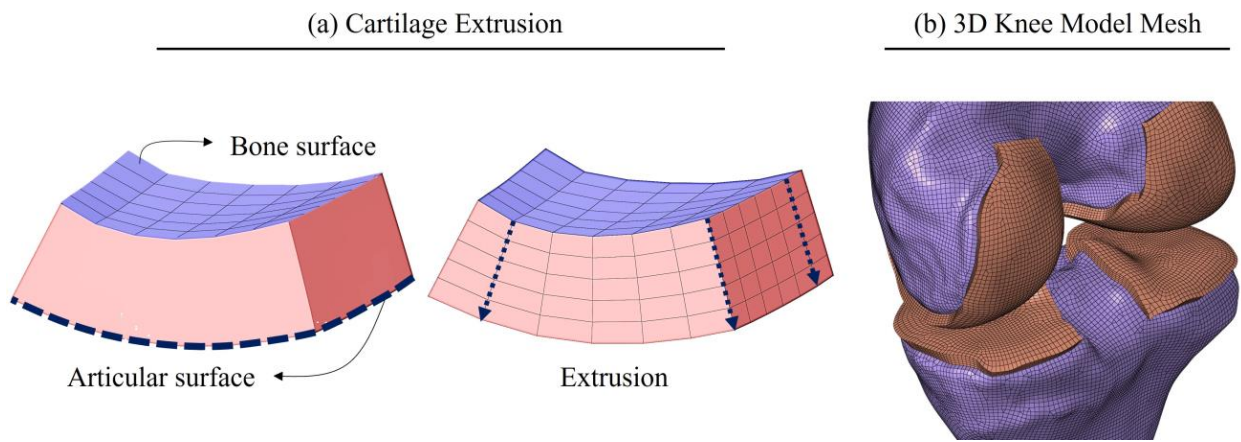

Figure 3. (a) Mapping the quadrilateral surface elements of the bone to the articular cartilage surface using solid elements. (b) 3D knee joint model

## 2.5 Computational modeling

Material and Finite element modeling

We employed the biphasic constitutive model proposed by Federico and Gasser [43] and Federico and Grillo [44] for cartilage, which consisted of an incompressible fluid phase and a fibril-reinforced solid/matrix phase. Collagen fibrils were separated into isotropic and anisotropic components under the assumption that the matrix was isotropic and inhomogeneous along its thickness. The directional orientation of the fibrillar network was captured by the anisotropic fibrils. *Table 1S in the supplementary material* provides a description of the biphasic model and associated material constants. The results of creep indentation experiments conducted by Pajerski [45] and Athanasiou et al. [46] on human knee cartilage served as the basis for the extracellular matrix (ECM) material properties.[47] A detailed explanation of the cartilage constitutive laws with its material constants used for describing cartilage behavior can be found in the *supplementary material* and also in our previous work [47], [48]. Briefly, the state of stress was defined by:

$$\sigma = -pI + \emptyset_0 \sigma_0 + \emptyset_1 (\sigma_{1i} + \sigma_{1a}) \quad \text{eq. 2}$$

where $\sigma$ is the total stress in the tissue, $p$ is the hydrostatic interstitial fluid pressure, $I$ is the unity tensor, and $\emptyset$ is the volume fraction. Here, subscripts 0 and 1 denote matrix and collagen fibrils, respectively. The matrix was considered isotropic, while the collagen fibrils were divided into isotropic ($\sigma_{1i}$) and anisotropic ($\sigma_{1a}$).

Menisci were assumed transversally isotropic and elastic, with elastic moduli of 20 MPa (for E1 and E2) and 159.6 MPa (for E3) [49]–[52]. To model the meniscal ligaments [53]–[56], linear elastic spring components with a stiffness of 375 N/mm per ligament were used. Elastic hexahedral elements (C3D8) and hexahedral pore pressure elements (C3D8P) were used to define the menisci and knee cartilage mesh, respectively. A surface-to-surface contact with frictionless tangential behavior was presented the contact mechanics of cartilage surfaces. Bones were considered as rigid bodies. The Anterior cruciate ligament (ACL), posterior cruciate ligament (PCL), medial and lateral collateral ligaments (MCL, LCL) were modelled as bi-linear springs that could withstand tension but not compression. Tensile strength k=380N/mm was used for the ACL [57], whereas k=200N/mm was used for the PCL [58]. Tensile stiffness for the MCL and LCL were k=100N/mm [58], [59].

The middle-central position between the medial and lateral epicondyles of the femur was used as the reference point for coupling the femur surface to loading [60], [61]. The end nodes of the meniscal horns and the bottom nodes of the tibial cartilage were fixed. The cartilage surfaces at the calcified zone were impermeable, while the pore pressure of the articular cartilage surfaces was set to zero, permitting free fluid flow. The loading (Mohammadi et al., 2020; Mononen et al., 2016) was applied at the reference point as a combination of an indentation load and a flexion moment [60], [62].

## 2.6 Statistical analysis

To compare the mechanical response from the manual and semi-automated FE models, 5 parameters, including the max principal stress, max principal strain, fluid pressure, fibril strain, and contact area, were considered for the duration of the stance simulation. The first 4 parameters were compared in superficial and deep zones, while the contact area was only calculated on the articular surface of the cartilage. In each zone, the average and peak values of these parameters were compared. We selected the statistical parametric mapping (SPM) method based on its inherent advantage in accommodating multiple comparisons when examining smooth and random 1-D trajectories. In contrast to traditional 0-D approaches such as the parametric t-test, the SPM method demonstrates superior suitability for this purpose [63]. The SPM t-test was performed for two independent samples with a criterion alpha-level of 0.05. The SPM was implemented using a Python package from https://spm1d.org/# for 1-D SPM.

## 3 Results

Table 1 presents the evaluation metrics for the segmentation performance of the femur and tibia using the Swin UNETR and SSM methods. The metrics include the DSC, Hausdorff distance, average distance, and the percentage of surface area associated with a distance greater than 1 mm. The DSC measures the overlap between the segmented regions and the ground truth (intersection over union). For all bone structures and segmentation methods, the DSC was consistently high, with a value over 98%. The Hausdorff distance quantifies the maximum distance from the nearest neighbor [64] between corresponding points on the segmented surface and the ground truth. The Swin UNETR method achieved a Hausdorff distance of 1.657±0.34 mm for femur and 1.65±0.48 mm for tibia. The SSM adjustment resulted in a slightly lower Hausdorff distance of 1.42±0.37 mm for femur and 1.47±0.41 mm for tibia. The other calculated parameter to assess the accuracy of the semi-automated method was the average distance, which represents the average separation between the segmented surface and the ground truth surface. For the femur and tibia, the Swin UNETR method resulted in an average distance of 0.3±0.04 mm and 0.31±0.03 mm, with the SSM adjustment yielding a slightly lower value of 0.23±0.05 mm and 0.25±0.043 mm, respectively. The percentage of surface area associated with a distance greater than 1 mm (Δarea%>1mm) represents the percentage of surface area where the distance between the segmented regions and the ground truth exceeds 1 mm. The femur and tibia segmentation using the Swin UNETR method showed a Δarea%>1mm of 0.98±1.61% and 1.11±1.25%, respectively. After the SSM adjustment, the Δarea%>1mm values slightly decreased to 0.57±1.1% and 0.71±1.01%, respectively. ***Figure 2S*** in the ***supplementary material*** illustrates a comparison of manual and Swin UNETR segmentations.

Statistical analysis showed no significant difference between the manual and semi-automated FE models for all 9 samples. Figure 4 depicts the SPM of maximum principal stress and strain, fluid pressure, fibril strain, and contact area as a function of time (s). All parameters were within the critical values indicating no significant difference (p-value>0.05).

The distribution of mechanical response over the surface and depth-wise at 20% and 80% of the stance phase were illustrated in Figure 5. The manual and semi-automated FE models resulted in a similar distribution of parameters. ***Figures 3S-6S*** in the ***supplementary material*** illustrates the distribution for each sample in tibial and femoral cartilages for each of the five mechanical responses.

The average and peak values of the mechanical parameters in the superficial and deep zones are illustrated in Figure 6. The dotted line represents the absolute differences between the two FE models. The contact region over the articular surface was projected into the deep zone to measure the mechanical parameters in the deep zone in Figure 6. The fluid pressure had the largest error of 0.01 MPa. In the ***supplementary material, Figures 7S – 10S*** are plotted for each sample separately to provide a more detailed comparison between the semi-automated and manual FE models. All these figures indicated no significant variation in the mechanical response of the semi-automated FE model compared to the manual FE model's.

Table 1. DSC, Hausdorff distance (mm), average distance (mm), and percentage of surface area associated with distance greater than 1 mm for Swin UNETR and SSM.

|  | DSC | Hausdorff distance (*mm*) | Average distance (*mm*) | $\Delta area\% > 1mm$ |
|---|---|---|---|---|
| Femur Swin UNETR | 98.51±0.09 | 1.66±0.34 | 0.30±0.04 | 0.98±1.61 |
| Tibia Swin UNETR | 98.59±0.06 | 1.65±0.48 | 0.31±0.03 | 1.11±1.25 |
| Femur SSM adjustment | 98.63±0.11 | 1.42±0.37 | 0.23±0.05 | 0.57±1.10 |
| Tibia SSM adjustment | 98.69±0.07 | 1.47±0.41 | 0.25±0.04 | 0.71±1.01 |

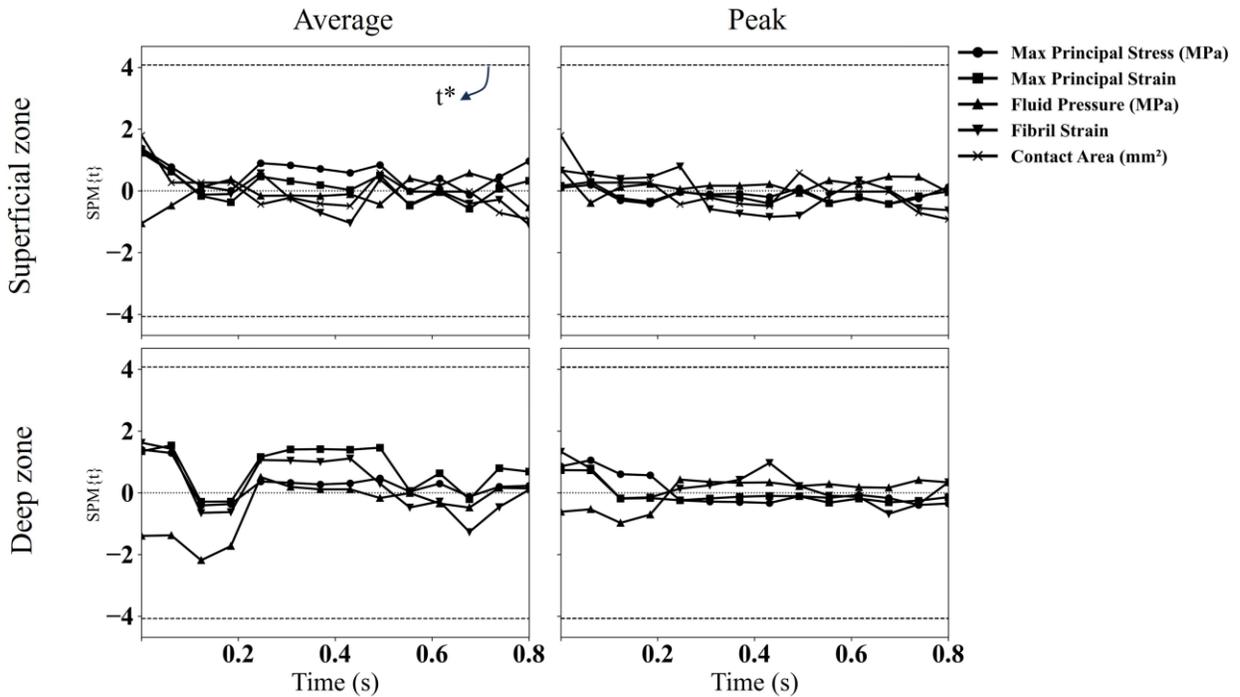

Figure 4. Statistical parametric mapping (SPM) as a function of time for the mechanical response of 5 parameters in superficial and deep zones. The dashed line shows the t-critical corresponding to a p-value of 0.05. The average column shows the average of the respected parameter over the contact region of all samples. Similarly, the Peak column shows the maximum value of these 5 parameters. The contact region over the articular surface was projected into the deep zone to calculate parameters in the deep zone.

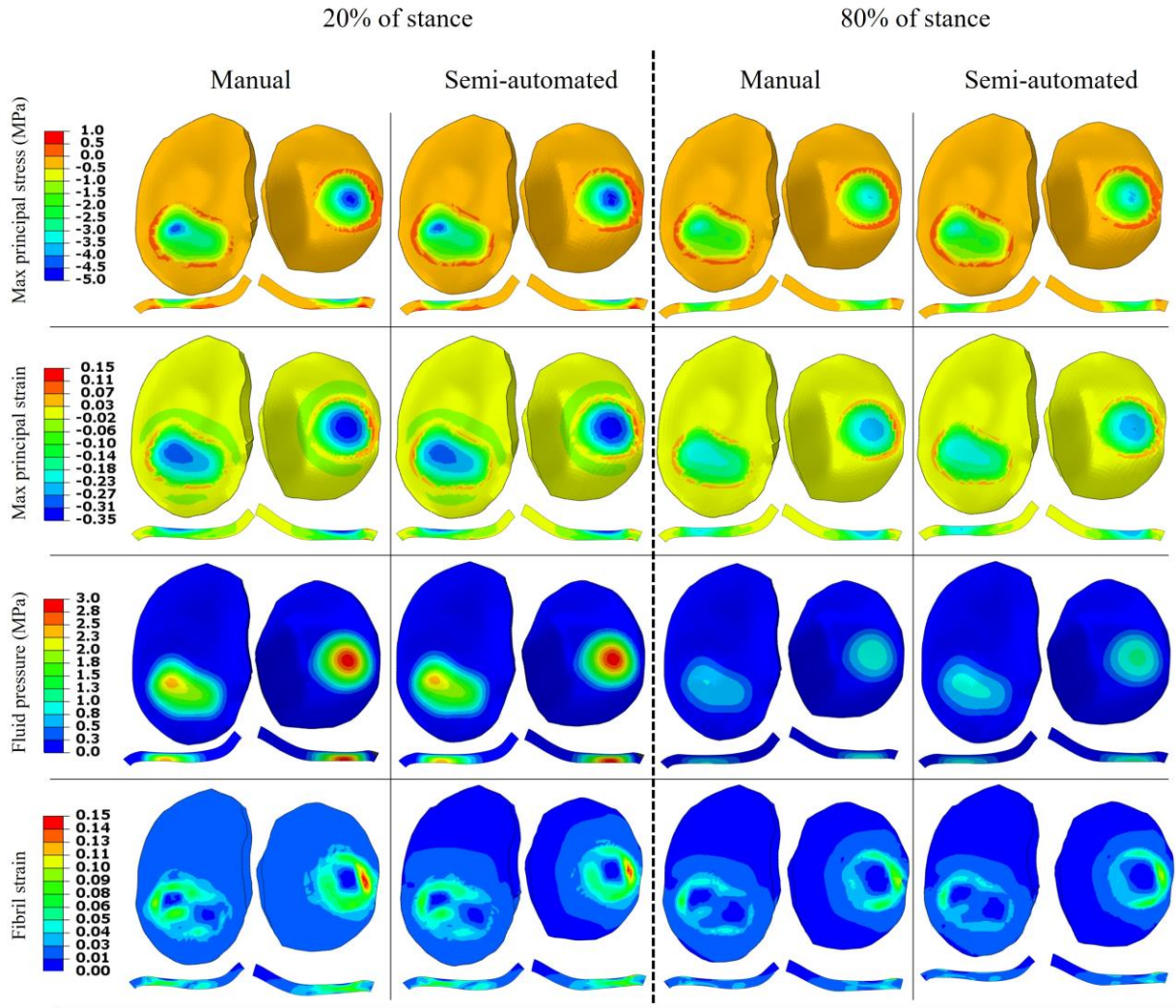

Figure 5. The distribution of maximum principal stress and strain, fluid pressure, and fibril strain over the surface and thickness of the tibial compartments at 20% and 80% of the stance phase. The depth-wise illustration was from the cross-section where the peak value occurred.

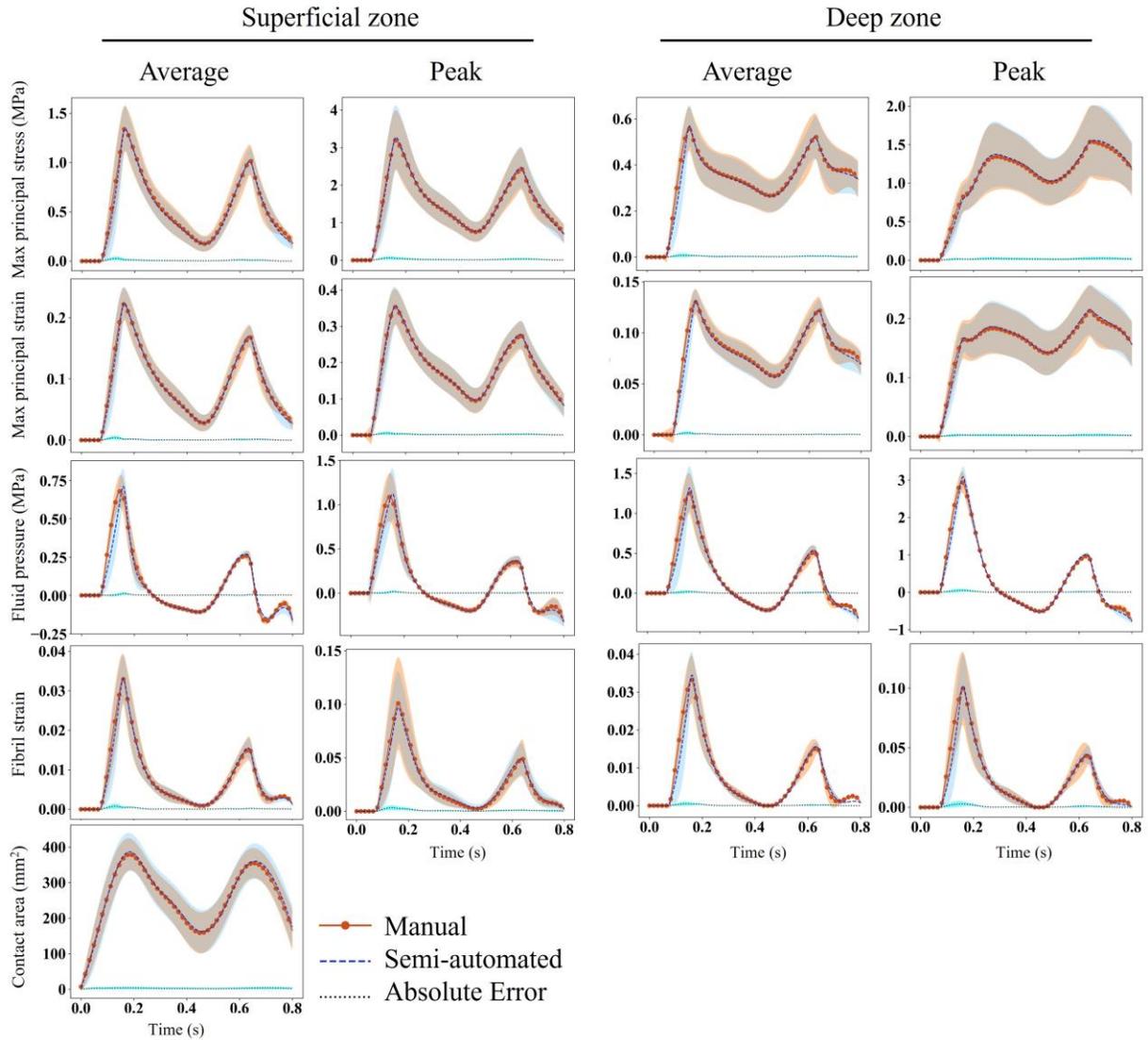

Figure 6. The average (over contact area) and maximum values of maximum principal stress and strain, fluid pressure, and fibril strain in the superficial and deep zones. The shaded region represents one standard deviation. The solid line with a circular marker represents the manual FE model, whereas the dashed line represents the semi-automated FE model. The dotted line is the absolute difference between the two models. The contact region over the articular surface of the cartilage was projected into the deep zone to calculate the parameters.

# 4  Discussion

In the present study, a trained SSM model of tibia and femur was mapped to a Swin UNETR segmentation model. The Swin UNETR helped developing a personalized model, and the SSM eliminated the postprocessing manual work needed to prepare the model for FE simulation, including filing holes and smoothing surfaces (Figure 7) [21], [24], [65]. By incorporating prior knowledge and capturing shape variations from a training dataset of 507 MRIs, the SSM adjustment consistently delivered high-quality surfaces in the context of image segmentation. These benefits make the proposed Swin UNETR and SSM a valuable semi-automated approach for accurate and robust FE model development from the tibia and femur MRIs.

Generally, geometrical models, such as SSM, require manual landmark selection by the user from medical images. This can negatively impact the accuracy, reproducibility, and segmentation time due to intra-individual variability. However, we tackled this challenge by employing Swin UNETR to generate unlimited anatomical landmarks automatically for SSM. Such models can capture spatial dependencies and long-range context information, leading to more precise segmentations [24].

Overall, the segmentation performance of the Swin UNETR model and SSM adjustment exhibited high DSC values (Table 1), indicating a strong agreement with the ground truth (i.e., the manual segmentation). The combination of Swin UNETR and SSM methods demonstrated lower Hausdorff distances and lower average distances compared to the Swin UNETR method, indicating better boundary conformity and closer agreement with the ground truth surface. Furthermore, the Δarea%>1mm values indicated minimal discrepancies in the segmented surface area for both methods (Table 1). In comparison to existing algorithms for segmenting knee images, DSC % achieved in our research are in the range of 98.63 for femur and 98.69 for tibia. These results are on par with the performance of other cutting-edge models. For instance, previous studies reported DSC % of 96.2 (tibia) and 97.0 (femur) [66], 98.5 (tibia) and 98.6 (femur) [24], 98.8 (tibia) and 98.6 (femur) [67], and 98.4 (femur and tibia) [23]. Nevertheless, it is important to acknowledge that making direct comparisons between these DSC scores across studies is challenging due to differences in training and test datasets typically used in each study [23].

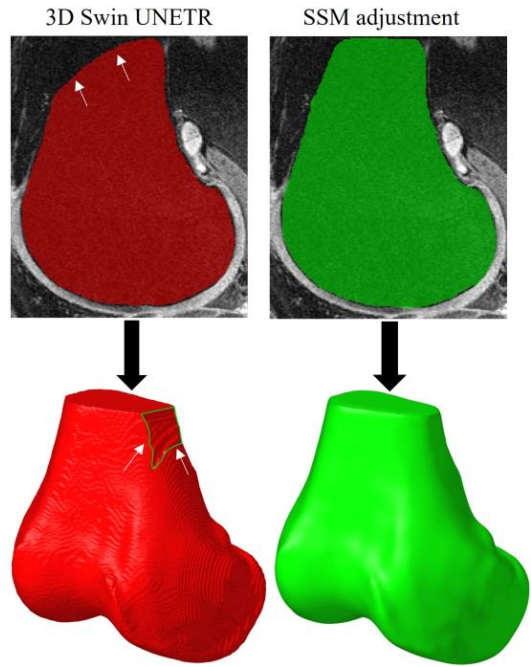

Figure 7. The outcome of the 3D Swin UNETR has a coarse surface topology with occasional artifacts, noises and holes (shown by arrows) that require post-processing before being used in a FE study. The SSM adjustment produced high-quality surfaces without compromising the accuracy of the segmentation.

The five parameters used for verifying the results of the semi-automated FE model in this study are the most common mechanical responses studied in knee joint biomechanics literature [14]–[16], [61], [62]. The 20% and 80% stance phase selected for evaluating parameters in Figure 5 corresponded to the two peaks of the loading condition (Figure 1). A strong agreement was found for the distribution of all parameters between the two models, except for fluid pressure, where the semi-automated FE model resulted in a larger fluid pressure at 80% stance phase compared to the manual FE model. This was reflected in a larger t-value of SPM for fluid pressure compared to the other four parameters (Figure 4); however, the respected values were well below the t-critical value and thus, the fluid pressure difference between the manual and semi-automated FE models was not significant.

Figure 5 illustrates a qualitative comparison of the mechanical responses between the manual and semi-automated FE models for one sample at 20% and 80% of the stance phase, while Figure 6 provides quantitative comparisons over the entire stance phase averaged for elements in the contact region, where the five parameters had higher magnitudes across the model. While some discrepancies were observed at the maximum values, it is noteworthy that they were not substantial, considering the absolute error indicated by the dotted line. The time scale in Figure 6 corresponds to the one used in statistical analysis (SPM) presented in Figure 4. From the analysis of these figures, it becomes evident that there were no significant differences (p-value >0.05) between the manual and semi-automated FE models across the entire stance phase and samples. The lack of significant differences and small absolute errors indicate that the semi-automated segmentation framework resulted in a knee FE model that predicts the same mechanical response as the manual segmentation. These results highlight the reliability and accuracy of the semi-automated segmentation approach, supporting its potential as a viable alternative to manual segmentation for the analysis of mechanical properties in the studied samples.

The current study has some limitations. One limitation of our study was the exclusion of the meniscus and cartilage contact in the finite element modeling of knee cartilage [68], [69]; however, in the context of our specific research objectives and scope, this omission does not significantly impact the findings and conclusions drawn. While the meniscus and cartilage contact play important roles in knee biomechanics, their inclusion would have significantly increased the computational time for FE modeling. Therefore, given the focus and objectives of our research, the decision to exclude these components does not compromise the validity and relevance of our study findings [16]. Another limitation of our study is the small number of samples used in FE modeling, which may limit the generalizability of our findings. We included nine samples for finite element modeling. A larger data set would enhance the generalizability of our results. Future studies should aim to include a larger sample size and consider the use of automatic meshing techniques [19], [70], while ensuring consistent and reliable geometries with computational runnability [63].

In summary, the integration of SWIN UNETR and SSM has demonstrated a remarkable effectiveness in the segmentation of MRIs. By harnessing the strengths of both SWIN UNETR and SSM, this method not only enhances segmentation precision but also creates suitable shapes and geometries for FE models. We have released our semi-automated segmentation models to the public ( https://data.mendeley.com/datasets/k5hdc9cz7w/1 ), aiming to contribute to the progress of biomechanical modeling and medical image segmentation. The ultimate goal of this study is to develop a segmentation of knee joint components, including cartilage, ligaments, and meniscus, to further facilitate computational modeling. This will help the biomechanical community swiftly achieving subject-specific knee joint segmentation.

**Acknowledgment:** The project was supported by the Natural Sciences and Engineering Research Council Canada (NSERC) Discovery grant [grant number 401610]; We would like to acknowledge and express our gratitude to Shirin Inanlou for her contribution to visual illustration and video prepration of supplementary material. Additionally, we would like to extend our appreciation to Alexander Tack for his assistance in the deep learning component. Their expertise and support greatly enhanced the quality of our research.

**Conflict of interest:** None

*Supplementary Material:*

# Integration of Swin UNETR and statistical shape modeling for a semi-automated segmentation of the knee and biomechanical modeling of articular cartilage

Reza Kakavand[1], Mehrdad Palizi[2], Peyman Tahghighi[1], Reza Ahmadi[1], Neha Gianchandani[1], Samer Adeeb[2], Roberto Souza[3,4], W. Brent Edwards[1], Amin Komeili[1*]

[1]Department of Biomedical Engineering, Schulich School of Engineering, University of Calgary

[2]Civil and Environmental Engineering Department, Faculty of Engineering, University of Alberta

[3]Department of Electrical and Software Engineering, Schulich School of Engineering, University of Calgary

[4]Hotchkiss Brain Institute, Cumming School of Medicine, University of Calgary

**Corresponding author: Amin Komeili**

**\* amin.komeili@ucalgary.ca**

**Address: ICT243, 2500 University Drive NW, Calgary, AB, T2N 1N4**

The matrix and fibril stresses were determined using energy functions:

$$\sigma = -pI + \emptyset_0 J^{-1} F \left(2\frac{\partial W_0(C)}{\partial C}\right) F^T + \emptyset_1 J^{-1} F \left(2\frac{\partial W_{1i}(C)}{\partial C} + 2\frac{\partial \bar{W}_{1a}(\bar{C})}{\partial C}\right) F^T \quad (2)$$

where $J$ is the determinant of $F$ (deformation gradient), $\bar{C}$ is the distortional component of the right Cauchy-Green deformation tensor, $C$. $W_0$ and $W_{1i}$ are the Holmes-Mow (Holmes and Mow, 1990) elastic strain energy potential of isotropic matrix and collagen fibrils, respectively, defined as:

$$W_{HM}(C) = \alpha_0 \frac{\exp\left[\alpha_1(I_1(C) - 3) + \alpha_2(I_2(C) - 3)\right]}{(I_3(C))^\beta} \quad (3)$$

Where $\alpha_0$, $\alpha_1$, $\alpha_2$ and $\beta$ are the material constants. $\bar{W}_{1a}$ was expressed as a function of the distortional component of $C$, because the collagen fibrils were assumed incompressible (Federico and Gasser, 2010; Komeili et al., 2020):

$$\bar{W}_{1a}(\bar{C}) = \int_{S_X^2} \psi(\vec{M}) \times \frac{1}{2} c_{1b} \left[\bar{I}_4\left(\bar{C}, A(\vec{M})\right) - 1\right]^2 dS \quad (4)$$

where $c_{1b}$ is a material constant, $\bar{I}_4$ is the fourth invariant of $\bar{C}$, $A(\vec{M}) = \vec{M} \otimes \vec{M}$ is a structure tensor that is a function of fibrils direction in the reference configuration $(\vec{M})$. The $\psi(\vec{M})$ is a probability distribution density function that gives the probability of finding a fibril aligned with the direction $\vec{M}$ (Federico and Gasser, 2010):

$$\psi(\vec{M}) = \rho(\Theta) = \frac{1}{\pi}\sqrt{\frac{b}{2\pi}} \frac{\exp[b(\cos 2\Theta) + 1]}{erfi(\sqrt{2b})} \quad (5)$$

Where $\Theta$ is the co-latitude in polar coordinates, $erfi(x)$ is the imaginary error function and the direction of the fibrils are controlled by the parameter b; negative and positive values of b produce parallel and perpendicular to surface fibrils orientation, while b=0 generates equally distribute fibrils over a sphere, i.e. random fibers distribution. Table 2 provides a description of the biphasic model and associated material constants.

Table 2. Material constants of the model (Komeili et al., 2020).

| Material properties | Collagen Fibril | ECM |
| --- | --- | --- |

|  | SZ | DZ | SZ | DZ |
|---|---|---|---|---|
| E (MPa)† | 10 | 15 | 2.5 | 3.8 |
| $v$ † | 0.3 | 0.3 | 0.1 | 0.1 |
| $\alpha_0$† | 3.4 | 5.1 | 0.6 | 1.0 |
| $\alpha_1$† | 0.1 | 0.1 | 0.8 | 0.8 |
| $\alpha_2$† | 0.4 | 0.4 | 0.1 | 0.1 |
| $c_{1b}$† (MPa) | 7.6 | 11.4 | – | – |
| $\beta$ † | 1.0 | 1.0 | 1.0 | 1.0 |
| $k$ † | – | – | 2.8 | 2.8 |
| $e_R$* † | – | – | 4.0 | 4.0 |
| b †† | 0 | 0 | 0 | 0 |
| Thickness‡ | 0.12h | 0.62h | 0.12h | 0.62h |

\* Void ratio (fluid / solid volume)
SZ: Superficial Zone
DZ: Deep Zone
h: Cartilage thickness
† (Pajerski, 2010)
†† (Komeili et al., 2020)
‡ (Julkunen et al., 2007)

In Figure 9, we present the results of the MRI segmentation comparison between the manual segmentation and the Swin UNETR segmentation method. The purpose of this evaluation is to visually assess the performance of the Swin UNETR model in segmenting MRI and to compare it against the gold standard of manual segmentation. The green contour overlaid on the Swin UNETR segmentation indicates the alignment with the outline obtained from manual segmentation.

The results from Figure 10, depicting the distribution of maximum principal stress over the surface and thickness of all nine samples at 20% of the stance phase, along with Figure 11, showing the distribution of maximum principal strain, both obtained from the cross-section where the peak value occurred, provided valuable insights into the mechanical behavior of the knee joint during loading conditions. Additionally, Figure 12 demonstrated the distribution of fluid pressure in the superficial zone and thickness of the nine samples at the same stance phase, while Figure 13 showcased the distribution of fibril strain.

The analysis of mechanical parameters in both the superficial and deep zones for all samples is depicted in Figure 14 to Figure 17. Figure 14 presents the average values of these parameters in the superficial zone, with the solid line representing the manual model and the dashed line denoting the automated model. The dotted line illustrates the absolute difference between the two models. Similarly, Figure 8 shows the peak values of the mechanical parameters in the superficial zone for all samples. It is important to note that the average values were calculated over the contact region, and the peak values reflect the peak values of the five parameters. Additionally, the contact region of the superficial zone was projected into the deep zone for calculating parameters in that region. Similarly, Figures 9 and 10 display the average and peak values of the mechanical parameters in the deep zone, respectively, for all samples.

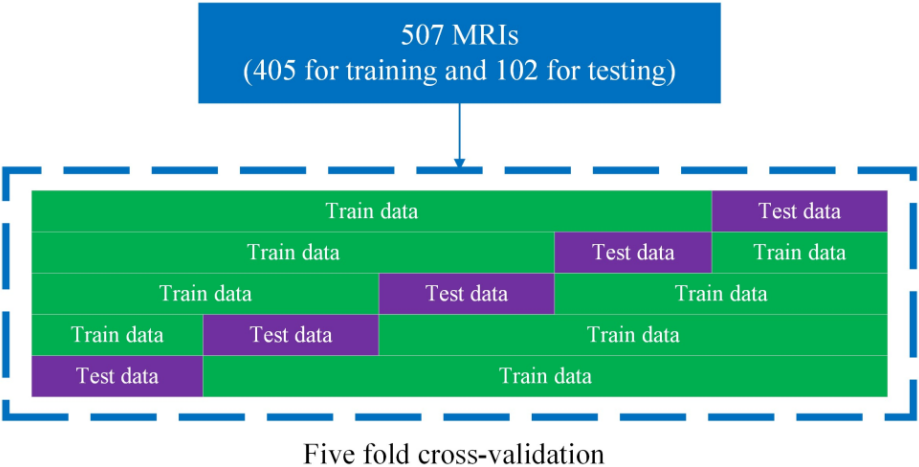

Figure 8. Five fold cross-validation used for training Swin UNETR and SSM models in our study.

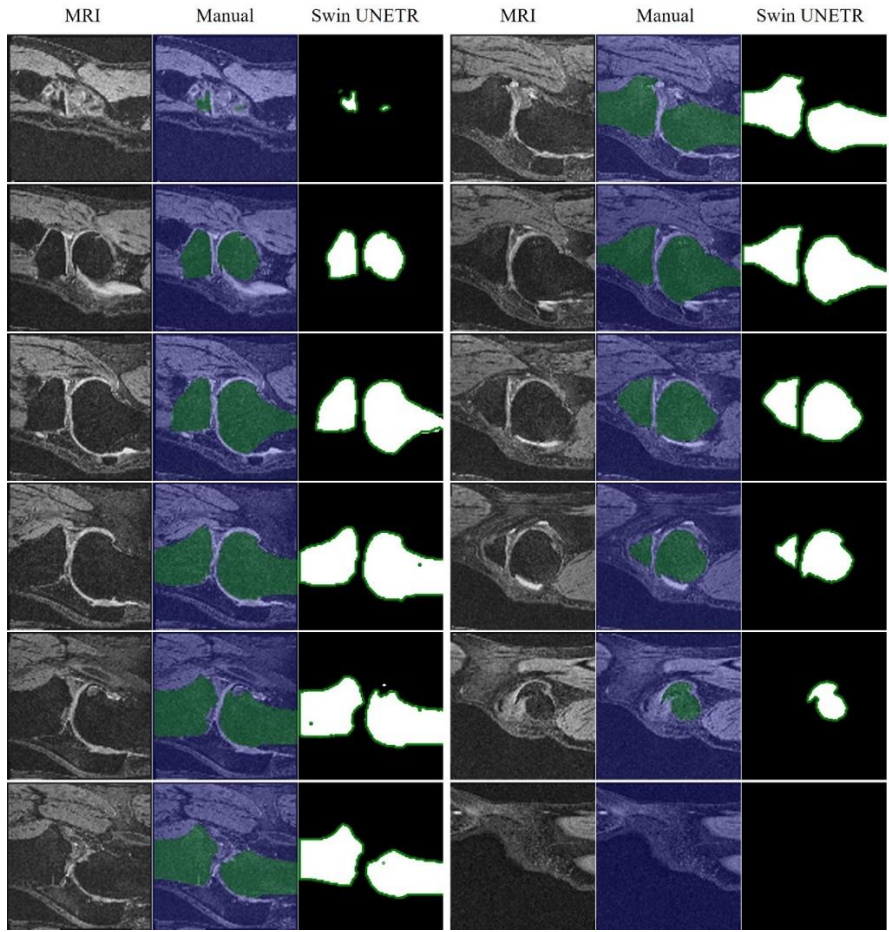

Figure 9. Example of MRI segmentation results comparison: manual segmentation and Swin UNETR segmentation. The green contour on Swin UNETR segmentation is the outline from the manual segmentation.

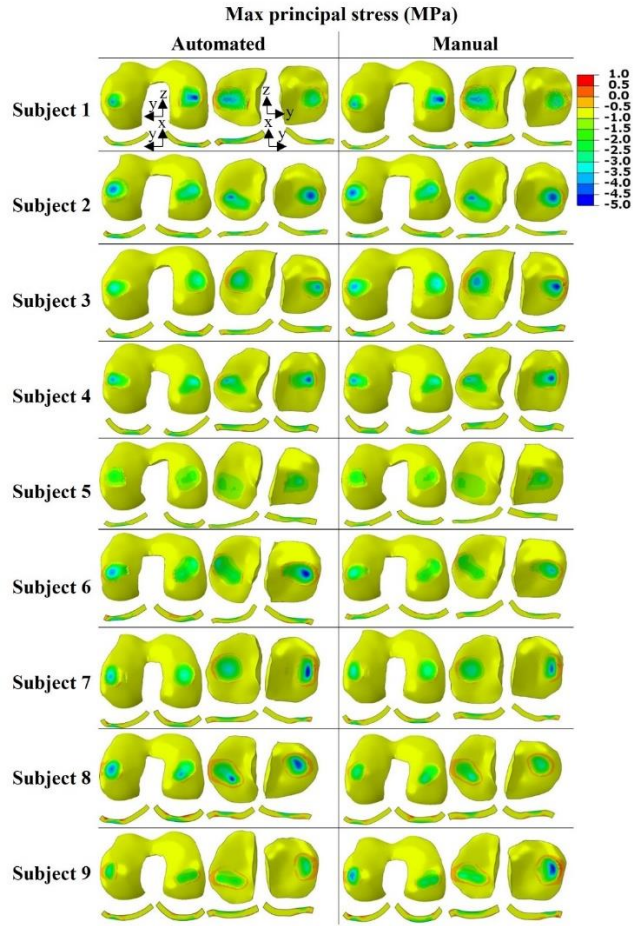

Figure 10. The distribution of maximum principal stress over the surface and thickness of all 9 samples at 20% of the stance phase. The depth-wise illustration was from the cross-section where the peak value occurred.

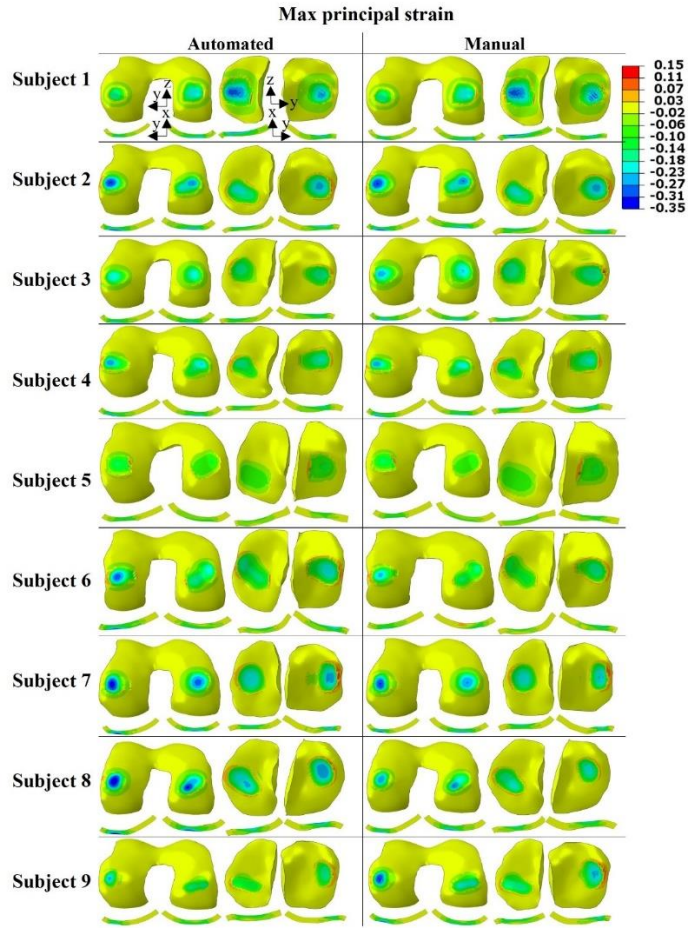

Figure 11. The distribution of maximum principal strain over the surface and thickness of all 9 samples at 20% of the stance phase. The depth-wise illustration was from the cross-section where the peak value occurred.

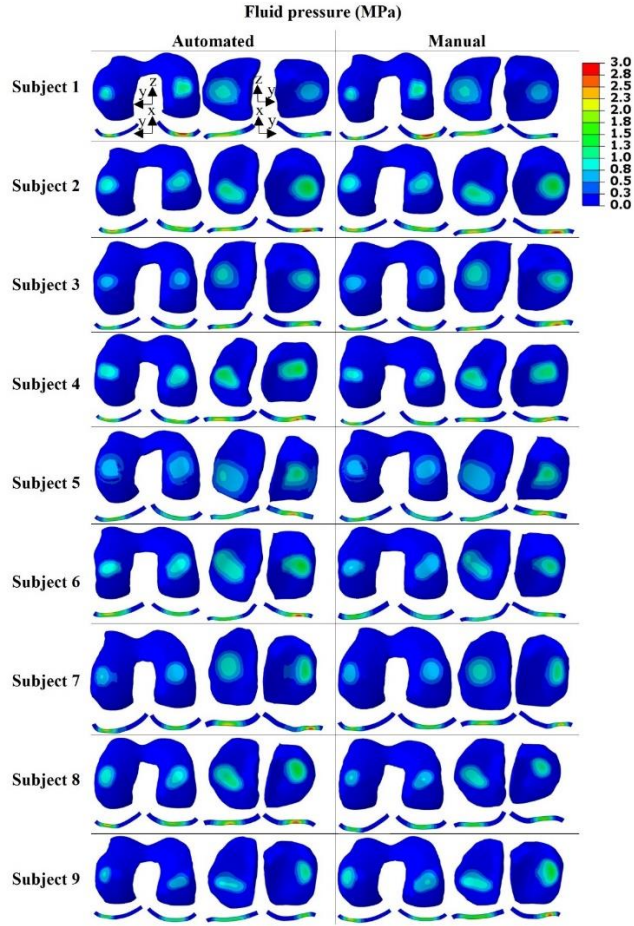

Figure 12. The distribution of fluid pressure in the superficial zone and thickness of all 9 samples at 20% of the stance phase. The depth-wise illustration was from the cross-section where the peak value occurred.

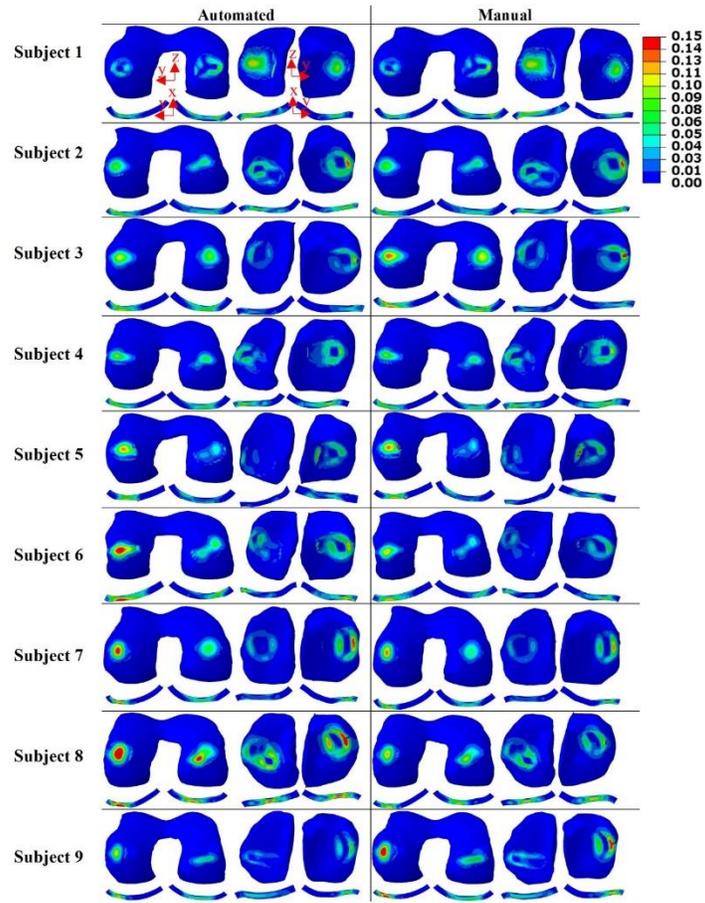

Figure 13. The distribution of fibril strain in over the surface and thickness of all 9 samples at 20% of the stance phase. The depth-wise illustration was from the cross-section where the peak value occurred.

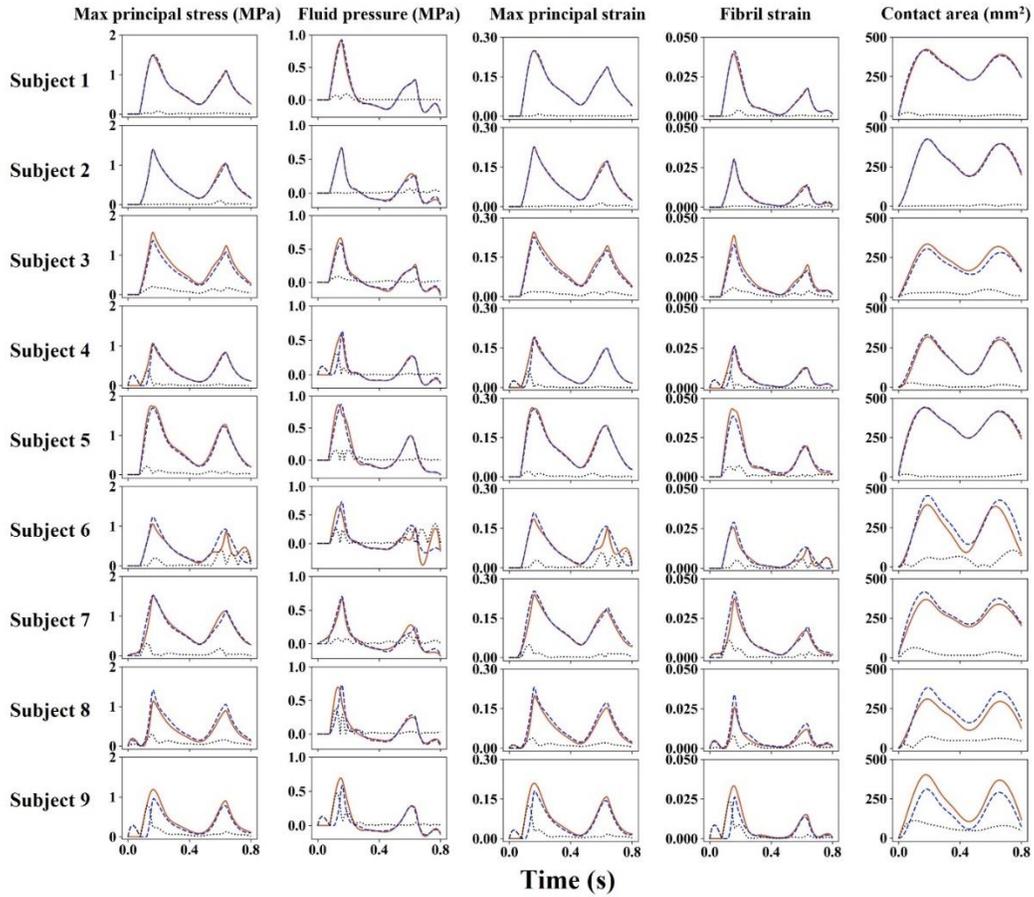

Figure 14. The **average** values of the mechanical parameters in **superficial zone** for all samples. The solid line was the manual model, whereas the dashed line was the automated model. The dotted line was the absolute difference between the two models. The average shows the average value over the contact region and peak shows the peak value of these 5 parameters. The contact region of the superficial zone was projected into the deep zone for calculating the parameters in the deep zone.

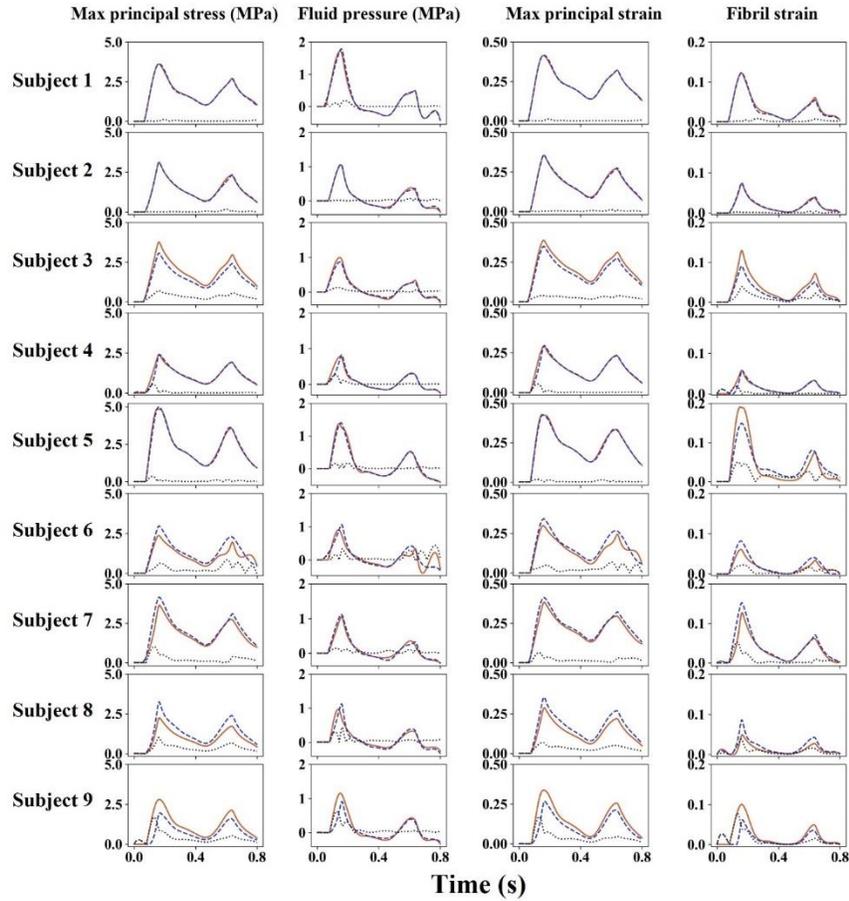

Figure 15. The **peak** values of the mechanical parameters in **superficial zone** for all samples. The solid line was the manual model, whereas the dashed line was the automated model. The dotted line was the absolute difference between the two models. The average shows the average value over the contact region and peak shows the peak value of these 5 parameters. The contact region of the superficial zone was projected into the deep zone for calculating the parameters in the deep zone.

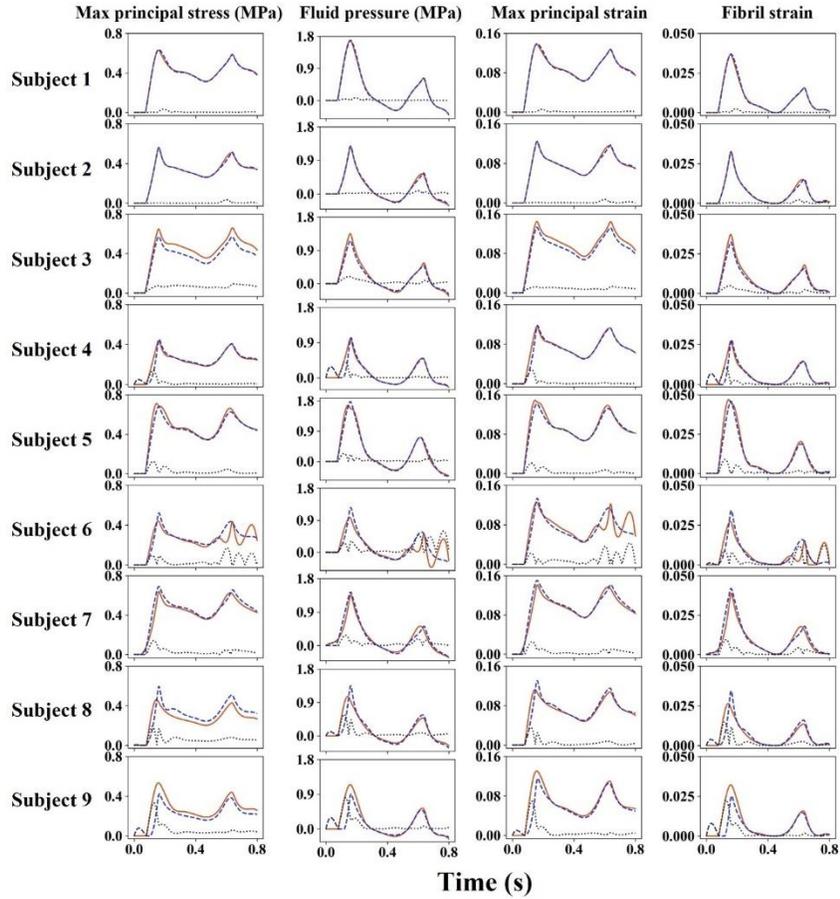

Figure 16. The **average** values of the mechanical parameters in **deep zone** for all samples. The solid line was the manual model, whereas the dashed line was the automated model. The dotted line was the absolute difference between the two models. The average shows the average value over the contact region and peak shows the peak value of these 5 parameters. The contact region of the superficial zone was projected into the deep zone for calculating the parameters in the deep zone.

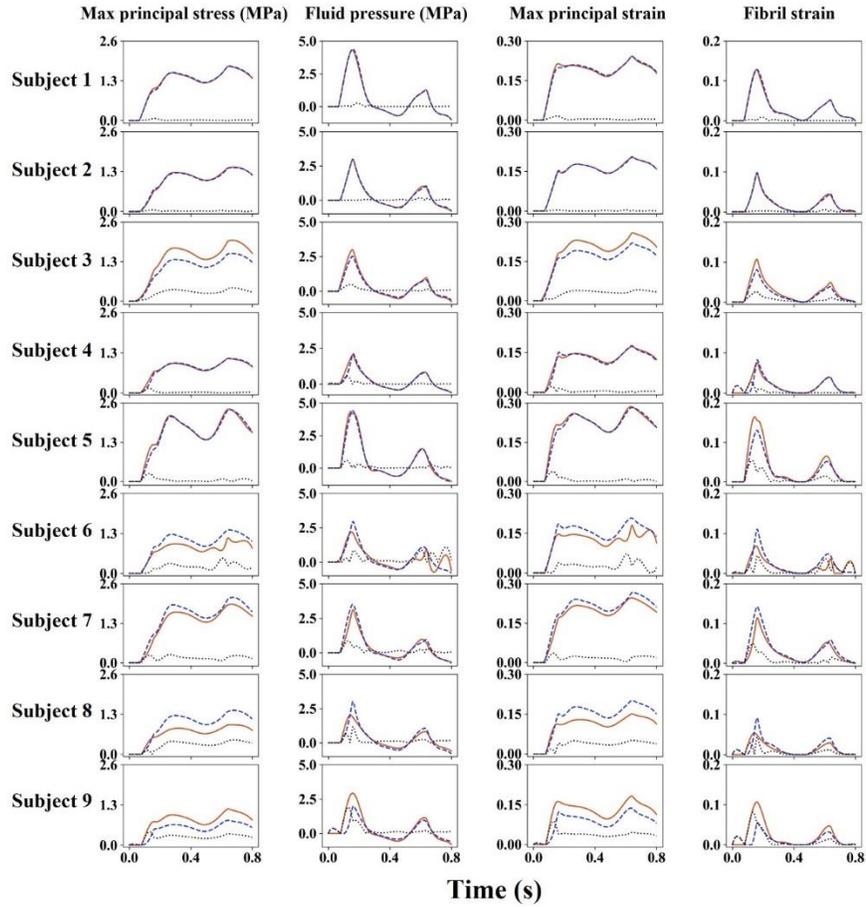

Figure 17. The **peak** values of the mechanical parameters in **deep zone** for all samples. The solid line was the manual model, whereas the dashed line was the automated model. The dotted line was the absolute difference between the two models. The average shows the average value over the contact region and peak shows the peak value of these 5 parameters. The contact region of the superficial zone was projected into the deep zone for calculating the parameters in the deep zone.